\pdfoutput=1

\documentclass[11pt]{article}
\usepackage{coling2020}
\usepackage{times}
\usepackage{url}
\usepackage{latexsym}
\usepackage{booktabs}
\usepackage{todonotes}
\newcommand{\projname}{\textsc{DaN+}}
\usepackage{floatrow}
\usepackage{paralist}
\usepackage{datetime}
\newdate{reddit}{28}{11}{2019}
\usepackage{tikz-dependency}
\usepackage{subcaption}
\usepackage{caption}
\usepackage{changes}

\colingfinalcopy

\title{\projname: Danish Nested Named Entities and Lexical Normalization}

 \author{Barbara Plank, Kristian Nørgaard Jensen and Rob van der Goot  \\
  Department of Computer Science\\
  ITU Copenhagen, Denmark \\
  {\tt bplank@itu.dk, krnj@itu.dk, robv@itu.dk} \\}

\date{}
\blfootnote{
    \hspace{-0.65cm}  
    This work is licensed under a Creative Commons 
    Attribution 4.0 International License.
    License details:
    \url{http://creativecommons.org/licenses/by/4.0/}.
}

\begin{document}
\maketitle
\begin{abstract}
This paper introduces \projname{}, a new multi-domain corpus and annotation guidelines for Danish nested named entities (NEs) and lexical normalization to support research on cross-lingual cross-domain learning for a less-resourced language. We empirically assess three strategies to model the two-layer Named Entity Recognition (NER) task. We compare transfer capabilities from German versus in-language annotation from scratch. We examine language-specific versus multilingual BERT, and study the effect of lexical normalization on NER. 
Our results show that
1) the most robust strategy is multi-task learning which is rivaled by multi-label decoding, 
2) BERT-based NER models are sensitive to domain shifts, and 
3) in-language BERT and lexical normalization are the most beneficial on the least canonical data.
Our results also show that an out-of-domain setup remains challenging, while performance on news plateaus quickly. This highlights the importance of cross-domain evaluation of cross-lingual transfer. 
\end{abstract}

\section{Introduction}

\begin{figure}[htb!]
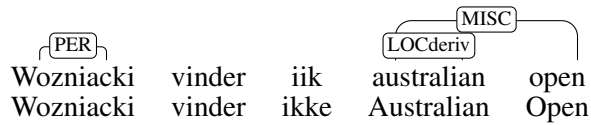

      \begin{dependency}[edge style={-},scale=0.5,distance=0.1em]
    \begin{deptext}[column sep=0.3cm]
    Wozniacki \& vinder \& iik \& australian \& open  \\
    Wozniacki \& vinder \& ikke \& Australian \& Open  \\
    \end{deptext}
    \depedge[edge height=2ex, edge slant=0pt, edge horizontal padding=25pt, edge end x offset=-25pt]{1}{1}{PER}
    \depedge[edge height=6ex, edge slant=0pt, edge end x offset=15pt, edge start x offset=-29pt]{4}{5}{MISC}
    \depedge[edge height=2ex, edge slant=0pt, edge horizontal padding=25pt, edge end x offset=-25pt]{4}{4}{LOCderiv}
    \end{dependency} 
\caption{Example with (a) nested entities and (b) lexical normalization. }
\label{fig:annotationexamples}
\end{figure}

Named Entity Recognition (NER) is the task of finding entities in text, such as locations, organizations, and persons. NER is a key step towards natural language understanding, for instance for question answering and information extraction. The task has received a substantial amount of attention, particularly for English. 
Most research so far, including for Danish, focused on newswire data and flat entities. It ignores  nested  entities, like `Australian Open' (illustrated in Figure~\ref{fig:annotationexamples}) being both an event and a location-derived entity. There is also little prior work on transfer learning for nested NER.

In this paper we introduce \projname{}, a novel resource for \textbf{Da}nish \textbf{N}ested \textbf{N}amed entities and lexical \textbf{N}ormalization, covering texts from canonical data from newswire and non-canonical social media sources. Danish bears interesting challenges for NER similar to German, which we capture by drawing inspiration from the \textsc{NoSta-D}~\cite{benikova2014nosta} NER annotation scheme. In particular, location adjectives like `dansk' (Danish) or `hollandske` (Dutch) are not capitalized, and there are tokens which are only partially named entities, like `Baltica-aktierne' (the Baltica shares). Such entities were mostly ignored so far. Full annotation guidelines for both tasks are provided in the appendix. 

\paragraph{Contributions} We present \begin{inparaenum}[1)]
\item \projname{}, a new multi-domain dataset for nested NER and lexical normalization;
\item  an evaluation of various models for Danish nested NER, including BERT variants and in-language versus cross-language experiments;
\item first experiments of lexical normalization on Danish and its downstream impact on NER.\footnote{All code and data to reproduce the experiments is available at
\url{https://github.com/bplank/DaNplus}} 
\end{inparaenum} 

\section{Related Work}\label{sec:relwork}

Nested NEs have received less research focus in contrast to flat entities~\cite{grishman-sundheim-1996-muc,grishman-1998-research,sang2003introduction,baldwin2015shared}. This has been attributed to technological complexity~\cite{finkel-manning-2009-nested} and limited  data availability~\cite{ringland-etal-2019-nne}. Existing nested NE data mostly spans newswire and biomedical data for English~\cite{kim2003genia,mitchell2005ace} and German news~\cite{benikova2014nosta}, for example.  Interest in nested NER is re-emerging~\cite{katiyar-cardie-2018-nested}, 
with many new recent neural approaches~\cite{sohrab2018deep,luan-etal-2019-general,lin-etal-2019-sequence,zheng-etal-2019-boundary}. To facilitate research, a fine-grained nested NER annotation on top of the Penn Treebank has been released recently~\cite{ringland-etal-2019-nne}.

To facilitate research on a less-resourced language, namely Danish, \newcite{plank-2019-neural} introduced publicly available evaluation data of flat NER on top of Danish UD~\cite{johannsen2015universal}, providing annotations for approximately 20\% of the data. The study also first benchmarked existing NER tools and evaluated the feasibility of transfer for Danish. 
~\newcite{hvingelby-EtAl:2020:LREC} recently independently annotated the entire Danish UD data for flat NERs, though with different guidelines, annotating also adjectives, for example.  Before these two recent studies, Danish NER data was behind a paywall or available tools were not benchmarked. 
To the best of our knowledge, \projname{} is the first Danish nested NER dataset beyond newswire.  

Domain shift is a pressing issue in NLP. One solution is to normalize the input text before detecting NEs, which is a mitigation strategy particularly suitable for social media~\cite{eisenstein-2013}. Previous work has evaluated lexical normalization for a variety of languages---but not for Danish---with  varying degrees of success~\cite{Schulz:2016:MTN:2906145.2850422,kucuk-steinberger-2014-experiments,nguyen2016text,liu2013named,li-liu-2015-improving,dugas-nichols-2016-deepnnner}. Most works do not evaluate the normalization model intrinsically, which is often restricted to a simple rule-based approach which is unlikely to transfer well.  \projname{}  provides also data to study lexical normalization for Danish.

To the best of our knowledge, there is very little prior work on cross-lingual and cross-domain transfer for nested (or overlapping) entities. Contemporary work includes English-Arabic~\cite{lan2020focused}.

\section{Data and Annotation}\label{sec:data}
 This section depicts the data sources and annotation.  Table~\ref{tab:overview} provides an overview of the \projname{} dataset. For normalization, as opposed to earlier annotation efforts in other languages, we included correction of capitalization. We refer to the appendix for details on annotation guidelines and data statement.
\subsection{Data varieties}

\projname{} includes canonical data from newswire and three social media varieties: 

\paragraph{News} The Danish DDT UD treebank~\cite{johannsen2015universal,buch-kromann}, which consists of news texts from PAROLE-DK~\cite{bilgram1998construction}. We use the canonical train/dev/test split.

\paragraph{Reddit} Sampled from the \texttt{r/Denmark} sub-reddit, in particular the top voted posts.\footnote{Using the universal Reddit scraper: \url{https://github.com/JosephLai241/Universal-Reddit-Scraper}} The collected posts all span a single date (November 28th 2019) and the data contains some non-Danish tokens (842 English tokens, 101 Swedish and 5 Norwegian).

\paragraph{Twitter}
We sample tweets collected over 2019-2020 using a list of Danish emotion words (love, pain, surprise), to avoid having mainly news articles. To make sure the data contains some phenomena interesting for normalization, we filtered it to contain at least 3 words not present in the Aspell dictionary.\footnote{We complemented the Aspell dictionary with some common named entities and interjections for this purpose.}

\paragraph{Arto}
Arto was Denmark's first large-scale social media platform and operated from 1988 till 2006. Because the website is not accessible anymore, we scraped all blog pages (where `blogs' can also consist of only a few words) and their corresponding comments from the Wayback Machine.\footnote{\url{https://archive.org/web/}} Similar to the Twitter data, we sample a subset and filter the data to contain some normalization density.

\begin{table}\centering
\resizebox{\textwidth}{!}{
\begin{tabular}{l|r|rrr|rr|rr|rr}
\toprule
Variety                   &  German: News & \multicolumn{3}{c}{\projname: News (UD-DDT)}  &  \multicolumn{2}{c}{Reddit} & \multicolumn{2}{c}{Twitter} & \multicolumn{2}{c}{Arto}\\
                   & \textsc{Train}  &\textsc{Train} &  \textsc{Dev} & \textsc{Test} & \textsc{Dev} & \textsc{Test} & \textsc{Dev} & \textsc{Test}  & \textsc{Dev} & \textsc{Test}  \\
                  \midrule
Sentences            & 24,002 & 4,383 & 564 & 565   & 326    &  126            & 120 & 110   & 336& 337\\
Tokens               & 452,853  & 80,378 & 10,332 & 10,023 & 4,547 & 4,497 & 5,347 & 5,086 & 5,496 & 4,389\\
Types                & 74,609 & 16,330 &  3,640  & 3,424  & 1,807 & 1,616 & 2,103 & 2,017 & 1,648 & 1,474\\
Sentences w/ NEs        &  59\% & 45\%  & 47\% & 48\% & 60\% & 56\%            & 80\% & 77\%   & 21\% & 20\% \\

1st level-NE        & 29,078 & 3,800 & 468 & 525  & 319  & 128             & 279 & 284     & 93 & 103\\
2nd level-NE        & 2,467   & 235   & 36 & 41     &  36  & 20              & 13 &  32      & 1  & 12\\
Tokens normalized    & --- & --- &  --- & --- & --- & --- & 3.5\% & 2.3\% & 16.7\% & 15.2\%\\
\bottomrule
\end{tabular}
}
\caption{Overview of \projname{}: \textbf{Da}nish \textbf{N}ested \textbf{N}amed entities and lexical \textbf{N}ormalization, which includes news and social media varieties (Reddit, Twitter, Arto). First column: GermEval~\cite{benikova2014nosta}. }
\label{tab:overview}
\end{table}

\subsection{Annotation}
We opted for a two-level NER annotation scheme following largely the annotation scheme provided by NoSTA-D~\cite{benikova2014nosta}.  \textit{First-level} annotations contain outermost entities (e.g., the company `Maribo Frø'). \textit{Second-level} annotations are sub-entities (location `Maribo').  
Four annotators were involved, three of which are native Danish speakers and one is proficient in Danish. For each task, a native speaker annotated the entire dataset after initial training. 
Inter-annotator agreement was high. For NER on the development sections of the Reddit and Twitter datasets, Cohen's $\kappa$ on the entity tokens without nesting was 90.97 and 83.08, respectively. With nesting, the $\kappa$ scores were 87.81 and 80.94. For lexical normalization, 10\% of the data was annotated by the two native speakers. For the decision on whether to normalize they reached a $\kappa$ of 88.66, whereas for the choice of the correct normalization the agreement was 96.30\%.

\section{Experimental Setup}

For nested NER, we use BERT~\cite{devlin-etal-2019-bert} with fine-tuning implemented in MaChAmp~\cite{machamp2020}.\footnote{MaChAmp v0.2, included in the repository.} 
We evaluate three decoding strategies: 
\begin{itemize}[-]
    \setlength\itemsep{.105cm}
    \item \texttt{single task-merged}: both annotation layers are merged into a single flat entity. 
    \item \texttt{multi-task}: the encoder is shared and each layer of annotation has its own decoder.
    \item \texttt{multi-label}: treats nested NER as multi-label problem, where a label $i$ is predicted if $P(l_i|\cdot) \geq \tau$~\cite{bekoulis2019neural} further illustrated in~\newcite{beesl}.

\end{itemize}

We first evaluate all NER models on Danish, both within news and on the three out-of-domain (OOD) varieties. We further compare to transfer from German: \begin{inparaenum}[1)] \item \texttt{zero-shot} transfer, fine-tuning only on German; and \item \texttt{union} of  the Danish and German data for fine-tuning. \end{inparaenum} We compare multilingual BERT (mlBERT)
versus training with Danish BERT (danishBERT).\footnote{version 2 from: \url{https://github.com/botxo/nordic_bert}} Even though both are trained on Danish data, for mlBERT this is Wikipedia data, whereas danishBERT is trained on Wikipedia, Common Crawl, Danish debate forums, and  Danish subtitles.
For MaChAmp, we use the proposed default parameters~\cite{machamp2020} shown to work well across tasks. We tune early stopping and $\tau$ on Danish news dev data, and set $\tau=0.9$. We compare our final model to the  \texttt{boundary-aware} model~\cite{zheng-etal-2019-boundary}, a state-of-the-art nested NER model which was also evaluated on GermEval 2014. We  train it with bilingual Danish and German Polyglot embeddings obtained via Procrustes alignment~\cite{conneau2017word}. For evaluation we use the official GermEval script~\cite{benikova2014nosta} with strict span-based F1 over both entity levels.\footnote{Compared to the evaluation of~\newcite{zheng-etal-2019-boundary}, this script is more strict. The scores we report are thus slightly lower compared to the ones reported in~\newcite{zheng-etal-2019-boundary}.}

For normalization, we choose to use MoNoise~\cite{van-der-goot-2019-monoise}, since it is open-source and is the only model that has shown to reach competitive performance across multiple languages. 
MoNoise requires n-grams and word embeddings to reach a good performance. We use a Wikipedia dump from 01-01-2020 and Twitter data collected throughout 2012 and 2018, filtered with the FastText language classifier~\cite{joulin2017bag}. 
Intrinsic normalization results are reported as capitalization sensitive word-level accuracy over all words (including words which are not normalized). Because we have no external training data for normalization, we use a 10-fold setup of dev+test.

\section{Results}
 
\begin{figure}
    \centering
    \includegraphics[width=\textwidth]{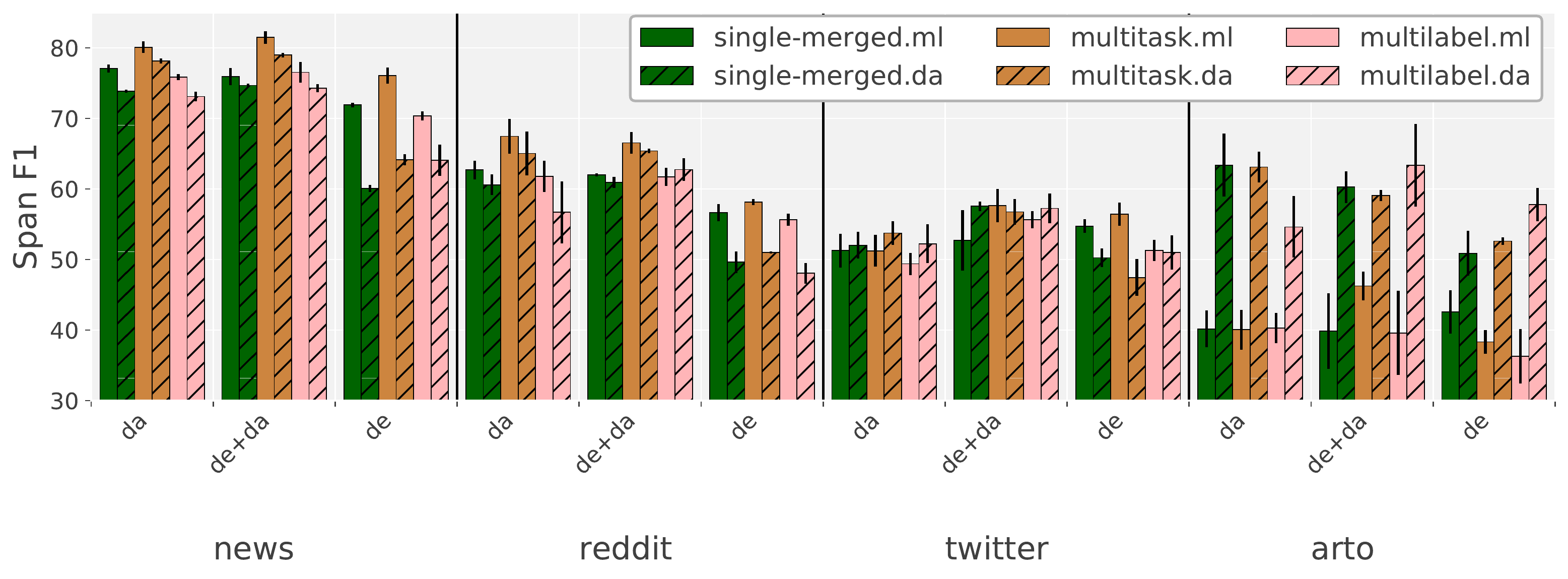}
    \caption{Nested NER results. Models trained on German (de), Danish (da) or both (de+da) and with mlBERT (ml) vs danishBERT (da). Average over 3 runs. Standard deviation indicated as  error line.}
    \label{fig:mainResults}
    
\end{figure}

\subsection{NER}
Figure~\ref{fig:mainResults} depicts the main results for nested NER on the dev set, while detailed results are given in Table~\ref{tab:fullResults} and~\ref{tab:nested} in the appendix.
First, we note that the model performs well within Danish newswire, reaching an F1 score in the 80ies (left bars). However, we observe a domain shift, as performances drop to 38-67\% on the three non-canonical social media datasets, with Twitter and Arto reaching lowest scores.

Our Danish training dataset is of modest size, hence the question arises whether  existing German data is beneficial. The German multi-task model performs remarkably well on Danish news in zero-shot setups with mlBERT, reaching an F1 of 76\%.  This can be explained by the closeness of the languages, the annotations and the large training data (Table~\ref{tab:overview}).

For a model trained on the union of the German and Danish data (\texttt{de+da}), we observe that
performance is overall close to the model trained on Danish only, which is five times smaller. The average F1 over all Danish datasets (News, Reddit, Twitter, Arto) for the two best models (using multi-task learning) is 65.01 with \texttt{da.da.multitask} and 65.05 with \texttt{de+da.da.multitask}. 
Interestingly, danishBERT is the best for the least non-canonical domain (Arto), in contrast to mlBERT which fares best on news. This is likely due to forum data included for pre-training danishBERT,\footnote{It should be noted that it is trained on lower-cased texts, which is suboptimal for NER yet works surprisingly well.} while mlBERT is based on Wikipedia data, which is less fit for non-canonical data.  This suggests that adaptive pre-training could yield better results~\cite{han_unsupervised_2019,ramponi-and-plank-2020-neural}.

We also compare transfer learning from German with increasing amounts of in-language Danish data. The learning curve in Figure~\ref{fig:learnc} shows that transfer helps for low amounts of data, and in-domain performance plateaus surprisingly quickly (especially for the \texttt{da+de} setup), and in-language data remains the best in-domain (ID). Instead, the gap to the non-canonical domains remains large for both in-language and cross-language setups, and performance on OOD is less stable throughout, calling for more out-of-domain evaluation of NER models.

\subsection{Lexical Normalization}

\begin{figure}

\begin{tikzpicture}
    \path[use as bounding box] (-9.25,-3) rectangle (6.25,3.1);

\node (word1) [] at (-5.45,  0) {
\begin{floatrow}
  \ffigbox[\FBwidth][]
    {\captionof{figure}{Learning curve for multi-task, mlBERT on in-domain (ID) news and average over all out-of-domain datasets (OOD).}
    \label{fig:learnc}}
    {\includegraphics[width=7cm]{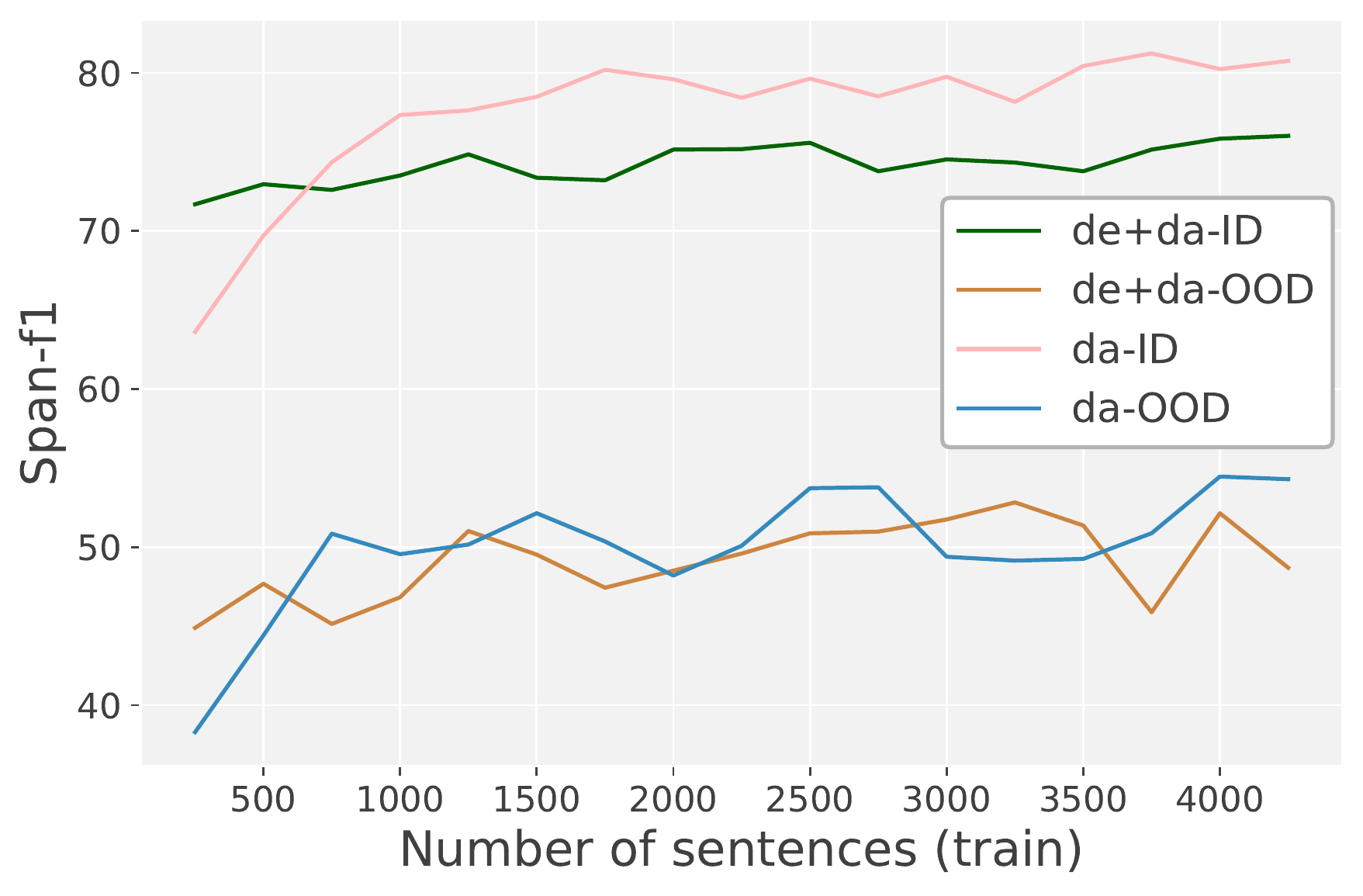}}
\end{floatrow}\hspace*{\columnsep}%
};

\node (word1) [] at (2.7,  -0.4) {
\begin{floatrow}
  \ffigbox[\FBwidth][]
    {\captionof{table}{Normalization accuracy, and its downstream effect on NER. For NER, multitask, mlBERT trained on de+da is used.}
    \label{tbl:normresults}}
    {
        \begin{tabular}{l  r r| r r}
        \toprule
               & \multicolumn{2}{c}{Normalization} & \multicolumn{2}{c}{NE tagging}\\
               & Twitter & Arto & Twitter & Arto \\
        \midrule 
        Baseline & 97.17 & 83.93 & 57.65 & 46.25 \\
        MoNoise & 97.17 & 92.52 & 58.59 & 55.83 \\
        Gold & 100.00 & 100.00 & 59.18 & 65.71 \\

        \bottomrule
        \\
        \\
        \end{tabular}

    }
\end{floatrow}\hspace*{\columnsep}%

};

\end{tikzpicture}
\end{figure}

We take a straightforward baseline for normalization, which always copies 
the original token, and we evaluate the impact of automatic versus gold normalization on NER. In other words, the accuracy of this baseline is equal to the percentage of not-normalized words. 
The results in Table~\ref{tbl:normresults} show that MoNoise is performing well for the less canonical Arto data in contrast to the Twitter data. On the Arto data, MoNoise reaches scores in a similar range compared to 
state-of-the-art results on other
languages~\cite{van-der-goot-2019-monoise}.\footnote{\newcite{van-der-goot-2019-monoise} 
use Error Reduction Rate (ERR) for evaluation, which is accuracy normalized for the amount
of words that need to be normalized; ERR in our setup would be
53.45,~\cite{van-der-goot-2019-monoise} report ERR's between 29 and 77.}

In the downstream evaluation (right part of Table~\ref{tbl:normresults}), we see that normalization is most beneficial when the data is less canonical (Arto), but even on Twitter normalization is beneficial. Furthermore, from the \textsc{gold} results, we can conclude that there is still space for improvement for automatic normalization.

\begin{table}

   \begin{tabular}{l r r r r r}
        \toprule
                        & German & News & Reddit & Twitter & Arto \\
        \midrule
        boundary-aware & 57.89 & 56.89 & 16.48 & 21.37 & 13.77\\ 
        \midrule
Raw (ml) & 83.31 & 80.73 & 57.99 & 60.87 & 54.88 \\
Norm'ed (ml) & --- & --- & --- & 61.74 & 56.38 \\
Raw (da) & 72.80 & 80.13 & 50.99 & 62.88 & 55.48 \\
Norm'ed (da) & --- & --- & --- & 62.91 & 56.59 \\

        \bottomrule
        \end{tabular}
    \caption{Nested NER F1 score on the test sets for models with mlBERT (ml) vs danishBERT (da).}
    \label{tab:test}

\end{table}

\subsection{Test Data}
We evaluate the model that fares overall best on in-domain source news (\texttt{de+da-multitask}) 
with danishBERT and mlBERT on the test sets.
Table~\ref{tab:test} shows that our model outperforms the \texttt{boundary-aware} method, which turns out to be brittle to domain shifts. Overall, the results confirm that normalization helps the most on the least canonical data (i.e.\ Arto), and  mlBERT is better than danishBERT on canonical news data, whereas on the least standard data (Arto) it is the other way around.

\section{Conclusions}
This paper contributes to the limited prior work on cross-lingual cross-domain transfer of nested NER. We provide a new resource for Danish, \projname{}, with baselines on nested NER and lexical normalization, using  two BERT variants and training on Danish, German or both. Our results show that BERT-based variants are sensitive to domain shift for cross-domain nested NER, whereas they can cope relatively well with missing in-language data. Results on normalization show that it helps in case of very non-standard data only, for which automatic normalization improves Danish nested NER performance.

\section*{Acknowledgments}
We thank Amanda Jørgensen for help with data annotation for lexical normalization.  We also thank NVIDIA, Google cloud computing and the ITU High-performance Computing cluster for computing resources.  This research is supported in part by the Independent Research Fund Denmark (DFF) grant 9131-00019A and  9063-00077B and an Amazon Faculty Research Award.

\bibliographystyle{coling}
\bibliography{biblio}

\begin{thebibliography}{}

\bibitem[\protect\citename{Baldwin \bgroup et al.\egroup
  }2015a]{lexnorm2015guidelines}
Timothy Baldwin, Marie-Catherine de~Marneffe, Bo~Han, Young-Bum Kim, Alan
  Ritter, and Wei Xu.
\newblock 2015a.
\newblock Guideline for {E}nglish lexical normalisation shared task.
\newblock Technical report, Workshop on Noisy User-generated Text.

\bibitem[\protect\citename{Baldwin \bgroup et al.\egroup
  }2015b]{baldwin2015shared}
Timothy Baldwin, Marie-Catherine de~Marneffe, Bo~Han, Young-Bum Kim, Alan
  Ritter, and Wei Xu.
\newblock 2015b.
\newblock Shared tasks of the 2015 workshop on noisy user-generated text:
  Twitter lexical normalization and named entity recognition.
\newblock In {\em Proceedings of the Workshop on Noisy User-generated Text},
  pages 126--135.

\bibitem[\protect\citename{Bekoulis}2019]{bekoulis2019neural}
Ioannis Bekoulis.
\newblock 2019.
\newblock {\em Neural Approaches to Sequence Labeling for Information
  Extraction}.
\newblock {Ph.D.} thesis, Ghent University.

\bibitem[\protect\citename{Bender and Friedman}2018]{bender-friedman-2018-data}
Emily~M. Bender and Batya Friedman.
\newblock 2018.
\newblock Data statements for natural language processing: Toward mitigating
  system bias and enabling better science.
\newblock {\em Transactions of the Association for Computational Linguistics},
  6:587--604.

\bibitem[\protect\citename{Benikova \bgroup et al.\egroup
  }2014]{benikova2014nosta}
Darina Benikova, Chris Biemann, and Marc Reznicek.
\newblock 2014.
\newblock Nosta-d named entity annotation for german: Guidelines and dataset.
\newblock In {\em LREC}, pages 2524--2531.

\bibitem[\protect\citename{Bilgram and Keson}1998]{bilgram1998construction}
Thomas Bilgram and Britt Keson.
\newblock 1998.
\newblock The construction of a tagged {D}anish corpus.
\newblock In {\em Proceedings of the 11th Nordic Conference of Computational
  Linguistics (NODALIDA 1998)}, pages 129--139.

\bibitem[\protect\citename{Conneau \bgroup et al.\egroup
  }2018]{conneau2017word}
Alexis Conneau, Guillaume Lample, Marc'Aurelio Ranzato, Ludovic Denoyer, and
  Herv{\'e} J{\'e}gou.
\newblock 2018.
\newblock Word translation without parallel data.
\newblock In {\em Sixth International Conference on Learning Representations}.

\bibitem[\protect\citename{Devlin \bgroup et al.\egroup
  }2019]{devlin-etal-2019-bert}
Jacob Devlin, Ming-Wei Chang, Kenton Lee, and Kristina Toutanova.
\newblock 2019.
\newblock {BERT}: Pre-training of deep bidirectional transformers for language
  understanding.
\newblock In {\em Proceedings of the 2019 Conference of the North {A}merican
  Chapter of the Association for Computational Linguistics: Human Language
  Technologies, Volume 1 (Long and Short Papers)}, pages 4171--4186,
  Minneapolis, Minnesota, June. Association for Computational Linguistics.

\bibitem[\protect\citename{Dugas and
  Nichols}2016]{dugas-nichols-2016-deepnnner}
Fabrice Dugas and Eric Nichols.
\newblock 2016.
\newblock {D}eep{NNNER}: Applying {BLSTM}-{CNN}s and extended lexicons to named
  entity recognition in tweets.
\newblock In {\em Proceedings of the 2nd Workshop on Noisy User-generated Text
  ({WNUT})}, pages 178--187, Osaka, Japan, December. The COLING 2016 Organizing
  Committee.

\bibitem[\protect\citename{Eisenstein}2013]{eisenstein-2013}
Jacob Eisenstein.
\newblock 2013.
\newblock What to do about bad language on the internet.
\newblock In {\em Proceedings of the 2013 Conference of the North {A}merican
  Chapter of the Association for Computational Linguistics: Human Language
  Technologies}, pages 359--369, Atlanta, Georgia, June. Association for
  Computational Linguistics.

\bibitem[\protect\citename{Finkel and Manning}2009]{finkel-manning-2009-nested}
Jenny~Rose Finkel and Christopher~D. Manning.
\newblock 2009.
\newblock Nested named entity recognition.
\newblock In {\em Proceedings of the 2009 Conference on Empirical Methods in
  Natural Language Processing}, pages 141--150, Singapore, August. Association
  for Computational Linguistics.

\bibitem[\protect\citename{Grishman and
  Sundheim}1996]{grishman-sundheim-1996-muc}
Ralph Grishman and Beth Sundheim.
\newblock 1996.
\newblock Message understanding conference- 6: A brief history.
\newblock In {\em {COLING} 1996 Volume 1: The 16th International Conference on
  Computational Linguistics}.

\bibitem[\protect\citename{Grishman}1998]{grishman-1998-research}
Ralph Grishman.
\newblock 1998.
\newblock Research in information extraction: 1996-98.
\newblock In {\em Proceedings of the {TIPSTER} Text Program: Phase {III}},
  pages 57--60, Baltimore, Maryland, USA, October. Association for
  Computational Linguistics.

\bibitem[\protect\citename{Han and Eisenstein}2019]{han_unsupervised_2019}
Xiaochuang Han and Jacob Eisenstein.
\newblock 2019.
\newblock Unsupervised domain adaptation of contextualized embeddings for
  sequence labeling.
\newblock In {\em Proceedings of the 2019 Conference on Empirical Methods in
  Natural Language Processing and the 9th International Joint Conference on
  Natural Language Processing (EMNLP-IJCNLP)}, pages 4238--4248, Hong Kong,
  China, November. Association for Computational Linguistics.

\bibitem[\protect\citename{Hvingelby \bgroup et al.\egroup
  }2020]{hvingelby-EtAl:2020:LREC}
Rasmus Hvingelby, Amalie~Brogaard Pauli, Maria Barrett, Christina Rosted,
  Lasse~Malm Lidegaard, and Anders Søgaard.
\newblock 2020.
\newblock Dane: A named entity resource for {D}anish.
\newblock In {\em Proceedings of The 12th Language Resources and Evaluation
  Conference}, pages 4597--4604, Marseille, France, May. European Language
  Resources Association.

\bibitem[\protect\citename{Johannsen \bgroup et al.\egroup
  }2015]{johannsen2015universal}
Anders Johannsen, H{\'e}ctor~Mart{\'\i}nez Alonso, and Barbara Plank.
\newblock 2015.
\newblock Universal dependencies for danish.
\newblock In {\em International Workshop on Treebanks and Linguistic Theories
  (TLT14)}, page 157.

\bibitem[\protect\citename{Joulin \bgroup et al.\egroup }2017]{joulin2017bag}
Armand Joulin, Edouard Grave, Piotr Bojanowski, and Tomas Mikolov.
\newblock 2017.
\newblock Bag of tricks for efficient text classification.
\newblock In {\em Proceedings of the 15th Conference of the European Chapter of
  the Association for Computational Linguistics: Volume 2, Short Papers}, pages
  427--431. Association for Computational Linguistics, April.

\bibitem[\protect\citename{Katiyar and Cardie}2018]{katiyar-cardie-2018-nested}
Arzoo Katiyar and Claire Cardie.
\newblock 2018.
\newblock Nested named entity recognition revisited.
\newblock In {\em Proceedings of the 2018 Conference of the North {A}merican
  Chapter of the Association for Computational Linguistics: Human Language
  Technologies, Volume 1 (Long Papers)}, pages 861--871, New Orleans,
  Louisiana, June. Association for Computational Linguistics.

\bibitem[\protect\citename{Kim \bgroup et al.\egroup }2003]{kim2003genia}
J-D Kim, Tomoko Ohta, Yuka Tateisi, and Jun’ichi Tsujii.
\newblock 2003.
\newblock Genia corpus—a semantically annotated corpus for bio-textmining.
\newblock {\em Bioinformatics}, 19(suppl\_1):i180--i182.

\bibitem[\protect\citename{Kromann \bgroup et al.\egroup }2003]{buch-kromann}
Matthias~T Kromann, Line Mikkelsen, and Stine~Kern Lynge.
\newblock 2003.
\newblock Danish dependency treebank.
\newblock In {\em International Workshop on Treebanks and Linguistic Theories
  (TLT)}, pages 217--220.

\bibitem[\protect\citename{K{\"u}{\c{c}}{\"u}k and
  Steinberger}2014]{kucuk-steinberger-2014-experiments}
Dilek K{\"u}{\c{c}}{\"u}k and Ralf Steinberger.
\newblock 2014.
\newblock Experiments to improve named entity recognition on {T}urkish tweets.
\newblock In {\em Proceedings of the 5th Workshop on Language Analysis for
  Social Media ({LASM})}, pages 71--78, Gothenburg, Sweden, April. Association
  for Computational Linguistics.

\bibitem[\protect\citename{Lan \bgroup et al.\egroup }2020]{lan2020focused}
Wuwei Lan, Yang Chen, Wei Xu, and Alan Ritter.
\newblock 2020.
\newblock A focused study to compare arabic pre-training models on newswire ie
  tasks.
\newblock In {\em arXiv 2004.14519}.

\bibitem[\protect\citename{Li and Liu}2015]{li-liu-2015-improving}
Chen Li and Yang Liu.
\newblock 2015.
\newblock Improving named entity recognition in tweets via detecting
  non-standard words.
\newblock In {\em Proceedings of the 53rd Annual Meeting of the Association for
  Computational Linguistics and the 7th International Joint Conference on
  Natural Language Processing (Volume 1: Long Papers)}, pages 929--938,
  Beijing, China, July. Association for Computational Linguistics.

\bibitem[\protect\citename{Lin \bgroup et al.\egroup
  }2019]{lin-etal-2019-sequence}
Hongyu Lin, Yaojie Lu, Xianpei Han, and Le~Sun.
\newblock 2019.
\newblock Sequence-to-nuggets: Nested entity mention detection via
  anchor-region networks.
\newblock In {\em Proceedings of the 57th Annual Meeting of the Association for
  Computational Linguistics}, pages 5182--5192, Florence, Italy, July.
  Association for Computational Linguistics.

\bibitem[\protect\citename{Liu \bgroup et al.\egroup }2013]{liu2013named}
Xiaohua Liu, Furu Wei, Shaodian Zhang, and Ming Zhou.
\newblock 2013.
\newblock Named entity recognition for tweets.
\newblock {\em ACM Transactions on Intelligent Systems and Technology (TIST)},
  4(1):1--15.

\bibitem[\protect\citename{Luan \bgroup et al.\egroup
  }2019]{luan-etal-2019-general}
Yi~Luan, Dave Wadden, Luheng He, Amy Shah, Mari Ostendorf, and Hannaneh
  Hajishirzi.
\newblock 2019.
\newblock A general framework for information extraction using dynamic span
  graphs.
\newblock In {\em Proceedings of the 2019 Conference of the North {A}merican
  Chapter of the Association for Computational Linguistics: Human Language
  Technologies, Volume 1 (Long and Short Papers)}, pages 3036--3046,
  Minneapolis, Minnesota, June. Association for Computational Linguistics.

\bibitem[\protect\citename{Mitchell \bgroup et al.\egroup
  }2005]{mitchell2005ace}
Alexis Mitchell, Stephanie Strassel, Shudong Huang, and Ramez Zakhary.
\newblock 2005.
\newblock Ace 2004 multilingual training corpus.
\newblock {\em Linguistic Data Consortium, Philadelphia}, 1:1--1.

\bibitem[\protect\citename{Nguyen \bgroup et al.\egroup }2016]{nguyen2016text}
Vu~H Nguyen, Hien~T Nguyen, and Vaclav Snasel.
\newblock 2016.
\newblock Text normalization for named entity recognition in {V}ietnamese
  tweets.
\newblock {\em Computational social networks}, 3(1):10.

\bibitem[\protect\citename{Plank}2019]{plank-2019-neural}
Barbara Plank.
\newblock 2019.
\newblock Neural cross-lingual transfer and limited annotated data for named
  entity recognition in {D}anish.
\newblock In {\em Proceedings of the 22nd Nordic Conference on Computational
  Linguistics}, pages 370--375, Turku, Finland, September{--}October.
  Link{\"o}ping University Electronic Press.

\bibitem[\protect\citename{Ramponi and
  Plank}2020]{ramponi-and-plank-2020-neural}
Alan Ramponi and Barbara Plank.
\newblock 2020.
\newblock Neural unsupervised domain adaptation in nlp---a survey.
\newblock In {\em COLING}.

\bibitem[\protect\citename{Ramponi \bgroup et al.\egroup }2020]{beesl}
Alan Ramponi, Rob van~der Goot, Rosario Lombardi, and Barbara Plank.
\newblock 2020.
\newblock Biomedical event extraction as sequence labeling.
\newblock In {\em Proceedings of the 2020 Conference on Empirical Methods in
  Natural Language Processing (EMNLP)}. Association for Computational
  Linguistics.

\bibitem[\protect\citename{Ringland \bgroup et al.\egroup
  }2019]{ringland-etal-2019-nne}
Nicky Ringland, Xiang Dai, Ben Hachey, Sarvnaz Karimi, Cecile Paris, and
  James~R. Curran.
\newblock 2019.
\newblock {NNE}: A dataset for nested named entity recognition in {E}nglish
  newswire.
\newblock In {\em Proceedings of the 57th Annual Meeting of the Association for
  Computational Linguistics}, pages 5176--5181, Florence, Italy, July.
  Association for Computational Linguistics.

\bibitem[\protect\citename{Schulz \bgroup et al.\egroup
  }2016]{Schulz:2016:MTN:2906145.2850422}
Sarah Schulz, Guy~De Pauw, Orph{\'e}e~De Clercq, Bart Desmet, V{\'e}ronique
  Hoste, Walter Daelemans, and Lieve Macken.
\newblock 2016.
\newblock Multimodular text normalization of {D}utch user-generated content.
\newblock {\em ACM Transactions on Intelligent Systems Technology}, 7(4):1--22,
  July.

\bibitem[\protect\citename{Sohrab and Miwa}2018]{sohrab2018deep}
Mohammad~Golam Sohrab and Makoto Miwa.
\newblock 2018.
\newblock Deep exhaustive model for nested named entity recognition.
\newblock In {\em Proceedings of the 2018 Conference on Empirical Methods in
  Natural Language Processing}, pages 2843--2849.

\bibitem[\protect\citename{Tjong Kim~Sang and
  De~Meulder}2003]{sang2003introduction}
Erik~F Tjong Kim~Sang and Fien De~Meulder.
\newblock 2003.
\newblock Introduction to the {C}o{NLL}-2003 shared task: Language-independent
  named entity recognition.
\newblock In {\em Proceedings of the Seventh Conference on Natural Language
  Learning at {HLT}-{NAACL} 2003}.

\bibitem[\protect\citename{van~der Goot \bgroup et al.\egroup
  }2020]{machamp2020}
Rob van~der Goot, Ahmet \"{U}st\"{u}n, Alan Ramponi, and Barbara Plank.
\newblock 2020.
\newblock {M}assive {C}hoice, {A}mple tasks ({MaChAmp}): A toolkit for
  multi-task learning in {NLP}.
\newblock {\em arXiv}.

\bibitem[\protect\citename{van~der Goot}2019]{van-der-goot-2019-monoise}
Rob van~der Goot.
\newblock 2019.
\newblock {M}o{N}oise: A multi-lingual and easy-to-use lexical normalization
  tool.
\newblock In {\em Proceedings of the 57th Annual Meeting of the Association for
  Computational Linguistics: System Demonstrations}, pages 201--206, Florence,
  Italy, July. Association for Computational Linguistics.

\bibitem[\protect\citename{Zheng \bgroup et al.\egroup
  }2019]{zheng-etal-2019-boundary}
Changmeng Zheng, Yi~Cai, Jingyun Xu, Ho-fung Leung, and Guandong Xu.
\newblock 2019.
\newblock A boundary-aware neural model for nested named entity recognition.
\newblock In {\em Proceedings of the 2019 Conference on Empirical Methods in
  Natural Language Processing and the 9th International Joint Conference on
  Natural Language Processing (EMNLP-IJCNLP)}, pages 357--366, Hong Kong,
  China, November. Association for Computational Linguistics.

\end{thebibliography}

\newpage

\appendix

\section{Full results}
Table~\ref{tab:fullResults} contains the exact scores which Figure~\ref{fig:mainResults} is based on. We also report the scores on only the nested entities in Table~\ref{tab:nested}; the multitask approach clearly
outperforms the other models for this category.

\begin{table}[h!]
    \begin{tabular}{l r r r r r}
    \toprule
  & German & News & Reddit & Twitter & Arto \\    
  \midrule
da.ml.single-merged & 66.32 & 77.09 & 62.71 & 51.26 & 40.16 \\
da.ml.multitask & 67.94 & 80.09 & \textbf{67.46} & 51.23 & 40.07 \\
da.ml.multilabel & 64.82 & 75.84 & 61.78 & 49.39 & 40.30 \\
da.da.single-merged & 30.35 & 73.85 & 60.59 & 52.02 & \textbf{63.37} \\
da.da.multitask & 33.94 & 78.15 & 65.04 & 53.75 & 63.11 \\
da.da.multilabel & 25.65 & 73.11 & 56.71 & 52.23 & 54.61 \\
de+da.ml.single-merged & 75.77 & 75.93 & 62.01 & 52.70 & 39.87 \\
de+da.ml.multitask & 83.88 & \textbf{81.48} & 66.53 & \textbf{57.65} & 46.25 \\
de+da.ml.multilabel & 76.36 & 76.53 & 61.72 & 55.62 & 39.58 \\
de+da.da.single-merged & 67.76 & 74.69 & 60.93 & 57.56 & 60.27 \\
de+da.da.multitask & 74.35 & 78.98 & 65.39 & 56.74 & 59.07 \\
de+da.da.multilabel & 66.47 & 74.31 & 62.75 & 57.25 & 63.35 \\
de.ml.single-merged & 75.50 & 71.90 & 56.63 & 54.74 & 42.59 \\
de.ml.multitask & \textbf{84.91} & 76.06 & 58.16 & 56.43 & 38.31 \\
de.ml.multilabel & 72.62 & 70.34 & 55.65 & 51.29 & 36.28 \\
de.da.single-merged & 67.21 & 60.08 & 49.62 & 50.25 & 50.87 \\
de.da.multitask & 74.20 & 64.14 & 51.01 & 47.45 & 52.61 \\
de.da.multilabel & 67.12 & 64.07 & 48.05 & 51.01 & 57.81 \\

\bottomrule
    \end{tabular}
    \caption{Span-f1 scores on all development sets for all out proposed models (single-merged, multi(task), multilabel), having two types of embeddings (da/ml), and all our training data combinations (da, de+da, de). Average over all Danish datasets (News, Reddit, Twitter, Arto) for the two best models are 65.01 for \texttt{da.da.multitask} and 65.05 for \texttt{de+da.da.multitask}. The latter is trained on 5 times more data while performing similarly to the model trained on Danish only.  }
    \label{tab:fullResults}
\end{table}

\begin{table}[h!]
    \begin{tabular}{l r r r r r}
    \toprule
  & German & News & Reddit & Twitter & Arto \\    
  \midrule
da.ml.single-merged & 4.15 & 10.94 & 4.94 & 3.17 & 0.00 \\
da.ml.multitask & 22.67 & 43.60 & 42.32 & 21.43 & 0.00 \\
da.ml.multilabel & 0.00 & 0.00 & 1.63 & 0.00 & 0.00 \\
da.da.single-merged & 2.30 & 10.38 & 5.31 & 0.00 & 0.00 \\
da.da.multitask & 6.30 & 39.44 & 47.23 & 22.16 & 0.00 \\
da.da.multilabel & 0.00 & 0.00 & 0.00 & 0.00 & 0.00 \\
de+da.ml.single-merged & 14.90 & 11.60 & 14.93 & 13.56 & 0.00 \\
de+da.ml.multitask & 65.80 & \textbf{47.61} & 39.10 & 15.74 & 0.00 \\
de+da.ml.multilabel & 2.56 & 0.00 & 0.00 & 0.00 & 0.00 \\
de+da.da.single-merged & 11.62 & 7.37 & 20.38 & 23.15 & 0.00 \\
de+da.da.multitask & 55.32 & 46.12 & \textbf{54.10} & \textbf{31.21} & 0.00 \\
de+da.da.multilabel & 0.84 & 0.00 & 0.00 & 0.00 & 0.00 \\
de.ml.single-merged & 15.77 & 8.20 & 18.98 & 19.17 & 0.00 \\
de.ml.multitask & \textbf{68.93} & 43.74 & 40.66 & 29.72 & 0.00 \\
de.ml.multilabel & 0.00 & 0.00 & 0.00 & 0.00 & 0.00 \\
de.da.single-merged & 11.88 & 7.14 & 17.19 & 21.60 & 0.00 \\
de.da.multitask & 56.42 & 30.21 & 25.93 & 24.02 & 0.00 \\
de.da.multilabel & 3.29 & 0.00 & 0.00 & 0.00 & 0.00 \\
\bottomrule
    \end{tabular}
    \caption{Span-f1 scores on all development sets for only the nested entities.}
    \label{tab:nested}
\end{table}

\clearpage
\section{\projname{} Data Statement}

Following~\cite{bender-friedman-2018-data}, the following outlines the data statement for \projname:

\textsc{A. CURATION RATIONALE} Collection of examples of Danish language for identification of named entities in different text domains, complemented with lexical normalization annotation to study the impact of it on NER. 

\textsc{B. LANGUAGE VARIETY} The non-canonical data was collected via the Twitter search API, the Reddit API and the Wayback archive.

Danish (da-DK) and some  US (en-US) mainstream English, Swedish (se-SE) and Norwegian (no-NO) in the Reddit sample.

\textsc{C. SPEAKER DEMOGRAPHIC} For the newswire data this is unknown. For the social media samples it is Danish and Scandinavian Reddit, Twitter and Arto users.  Gender, age, race-ethnicity, socioeconomic status are unknown. 

\textsc{D. ANNOTATOR DEMOGRAPHIC} Three students and one faculty (age range: 25-40),  gender: male and female. White European. Native language: Danish, German. Socioeconomic status: higher-education student and university faculty.

\textsc{D. SPEECH SITUATION} Both standard and colloquial Danish, i.e., edited and spontaneous speech. Time frame of data between 1988 and 2020.

\textsc{D. TEXT CHARACTERISTICS}  Sentences from journalistic edited articles and from  social media discussions and postings.

\textsc{PROVENANCE APPENDIX}  The news data originates from the Danish UD DDT data, GNU Public License, version 2 OR CC BY-SA 4.0: \url{https://github.com/UniversalDependencies/UD_Danish-DDT/blob/master/README.md}

\clearpage
\section{Annotation guidelines for NER}

This section describes the  annotation guidelines which we used for our \projname{} NER corpus. Our guidelines were adopted from the German NoSta-D guidelines~\cite{benikova2014nosta}.

We stick to a two layer annotation, where the outermost embraces the longer span and is the most prominent entity reading, and the inner span contains secondary or sub-entity readings. If there would be more than 2 layers, we drop the second potential reading in favor of keeping two layers (e.g., Australian Open is both an event and hence MISC but also an ORG; as Australian is a LOCderiv, we here keep only MISC for the event and LOCderiv for Australian).

\paragraph{Step 1:} Named entities are nominal phrases that determine specific people, organizations, locations or miscellaneous specific objects like film titles or products. National holidays or religious events (\textit{Jul, Ramadan}) are not annotated. Given the following example:

$$[\mbox{Leila}] \mbox{ bought } [\mbox{the house}]$$

There are two nominal phrases. Only one of them is a named entity (Leila), the second nominal is a common noun. 

\paragraph{Step 2: Potential NEs} Only full nominal phrases are potential full NEs. Pronouns and all other phrases should be ignored. Derivations of NEs, i.e., words which are derived through morphological derivation processes, are marked (e.g., \textit{danske}). NEderiv do not need to be nominal phrases. Declination (e.g., genitive forms) are not considered derivations and are directly annotated as NEs. For mediums such as social media we do mark user names and hashtags as potential NEs. Note that we diverge from the German NoSta-D guidelines by annotating the names of languages (e.g., \textit{dansk, swahili}) as LOCderiv. 

\begin{itemize}
    \item Full NEs are annotated as LOC (location), ORG (organization), PER (person) or MISC (miscellaneous other)
    \item Derivations of NEs are marked as such by appending \textit{deriv}, e.g., den [danske]\textit{LOCderiv} midtbanespiller
\end{itemize}

Examples:
\begin{itemize}
    \item Location: 
    \textit{[København]LOC},  \textit{[Kastrup]LOC}
    \item BUT when the location acts as an organized entity (e.g. country, municipality, sports club), it is tagged as ORG with LOC as inner layer: \textit{[[Danmark]LOC]ORG indfører grænsekontrol}
    \item \textit{[Carsten Jensen]PER}
    \item \textit{[IKEA]ORG}
    \item \textit{[Parken]LOC} (Stadium) 
    \item \textit{[The Shining]MISC}, \textit{[Jojo]MISC} (product name, song titles etc)
    \item Location adjectives: 
    \textit{De [københavnske]LOCderiv gader}
    \item Person adjectives:
    \textit{[Freudiansk]PERderiv litteratur}
    \item BUT genitive forms: \textit{[[Denmarks]LOC Radio]ORG},  \textit{[Københavns]LOC kommune}, \textit{[Johannsons]PER hus}
\end{itemize}

Examples:
\begin{itemize}
    \item Organizations: 
    \textit{[Twitter]ORG},  \textit{[TV2]ORG}, på min [FB]ORG
    \item BUT: 
    reference to specific Reddit channels \textit{[\//r\//all]MISC}
    \item \textit{at være [dansker]LOCderiv på [reddit]ORG}
    
\end{itemize}

\paragraph{Step 3: Titles, owners} Determiners and titles are not part of NEs. But owners can be NEs by itself.

Examples: 
\begin{itemize}
    \item 
    \textit{dronning [Margareth]PER}, \textit{dronning [Margareth II]PER} (numbers are kept as part of the name)
    \item \textit{[Vivaldis]PER [Vier Jahreszeiten]MISC}
\end{itemize}

\paragraph{Step 4: Multi-word tokens} NEs often consist of multiple tokens.

Examples: 
\begin{itemize}
    \item person names:
    \textit{[Terry Hatcher]PER}
    \item film titles (MISC):  \textit{[Breaking Bad]MISC}
\end{itemize}

\paragraph{Step 5: Nesting} NEs can be nested.

Examples: 
\begin{itemize}
    \item locations in organization names:
    \textit{[[Allerød]LOC Gymnasium]ORG \\ ([[Nordjyllands]LOC politi]ORG)} 
    \textit{}
    \item organization names in product names:
    \textit{[[Google]ORG Translate]MISC}
\end{itemize}

\paragraph{Step 6: Parts} Named entities can also be parts of tokens and are annotated as such with the suffix ``part".

Examples: 
\begin{itemize}
    \item 
    \textit{[pro-hongkong]LOCpart}
    \item \textit{[Hverdags-Lars]PERpart}
\end{itemize}

\paragraph{Step 6: Medium-specific potential NEs} Named entities can also be parts of special medium-specific tokens, like user names and hashtags in Twitter. We do annotate them as such.

Examples: 
\begin{itemize}
    \item 
    \textit{[@hik\_fodbold]ORG}
    \item \textit{[\#ToppenAfPoppen]MISC} 
    \item \textit{[@realDonaldTrump]PER}
\end{itemize}

\clearpage
\section{Annotation guidelines for lexical normalization}
The guidelines are based on~\cite{lexnorm2015guidelines}, all the cases were we diverged from these guidelines, or when we believed clarification was necessary are described below.

\textbf{Systematic miss-spellings} \newline
Since the data was taken from social media some words were systematically spelled wrong. This is especially seen on Arto, where many words were spelled using q instead of g. Here q was replaced with g:
\vspace{.2cm}

\begin{tabular}{l l}
jeq $\mapsto$ jeg (I) & muliqe $\mapsto$ mulige  \\
\end{tabular}
\vspace{.2cm} \newline

As it is also common to write words without the last one or two letters, there were also many words missing one or multiple letters in the end. Here the missing letters were inserted: 

\vspace{.2cm}

\begin{tabular}{l l}
ik $\mapsto$ ikke & hva $\mapsto$ hvad  \\
\end{tabular}
\vspace{.2cm} \newline

\textbf{Capitalization} \newline
Capitalization was corrected in names, first letter in a post and after periods, question marks and other signs that require capitalization in the first letter of the following word.
Capitalized words that illustrate yelling or emphasis have been decapitalized, acronyms that are capitalized have been kept capitalized. 

\vspace{.2cm}

\begin{tabular}{l l}
TILLYKKE $\mapsto$ tillykke (congratulations) & DR $\mapsto$ DR (Denmark's Radio)  \\
\end{tabular}
\vspace{.2cm} \newline

\textbf{Splitting and merging} \newline
Words that were incorrectly split or incorrectly merged into one word were corrected. 

\vspace{.2cm}

\begin{tabular}{l l}
ar bej der $\mapsto$ arbejder (works) & istedet $\mapsto$ i stedet (instead)  \\
\end{tabular}
\vspace{.2cm} \newline

\textbf{Phrasal abbreviations} \newline
There was no correction of phrasal abbreviations because the written-out form does not correspond to the intended meaning of the phrase. The only ones found were in English.

\vspace{.2cm}

\begin{tabular}{l l}
lol $\mapsto$ lol & omg $\mapsto$ omg  \\
\end{tabular}
\vspace{.2cm} \newline

\textbf{Hashtags} \newline
Hashtags and usernames were not corrected, even if they were misspelled or if they contained multiple words.
\vspace{.2cm}

\begin{tabular}{l l}
\#sundhedforalle $\mapsto$ \#sundhedforalle (health for all)  \\
\end{tabular}
\vspace{.2cm} \newline

\textbf{Corrections of the letters æ, ø, å} \newline
The Danish alphabet contains the three letters æ, ø and å. If these are not available at the used keyboard they are often replaced by other vowels: 
\vspace{.2cm}

\begin{tabular}{l l l}
ae $\mapsto$ æ & o $\mapsto$ ø & aa $\mapsto$ å  \
\end{tabular}
\vspace{.2cm} \newline

In words where the replacement vowels are used they have been replaced with the appropriate letter. In some data æ, ø and å were left out entirely, here the letters were inserted. As the missing letter in some cases results in multiple options, the word was determined using the context: 

\vspace{.2cm}

\begin{tabular}{l l}
har $\mapsto$ har (to have) or hår (hair) & fler $\mapsto$ flere (more) or føler (feels)  \\
\end{tabular}
\vspace{.2cm} \newline

\end{document}